# Prediction in ungauged regions with sparse flow duration curves and input-selection ensemble modeling

Dapeng Feng, Kathryn Lawson and Chaopeng Shen[1]
Civil and Environmental Engineering, Pennsylvania State University

**Abstract**

While long short-term memory (LSTM) models have demonstrated stellar performance with streamflow predictions, there are major risks in applying these models in contiguous regions with no gauges, or predictions in ungauged regions (PUR) problems. However, "softer" data such as the flow duration curve (FDC) may be already available from nearby stations, or may become available. Here we demonstrate that sparse FDC data can be migrated and assimilated by an LSTM-based network, via an encoder. A stringent region-based holdout test showed a median Kling-Gupta efficiency (KGE) of 0.62 for a US dataset, substantially higher than previous state-of-the-art global-scale ungauged basin tests. The baseline model without FDC was already competitive (median KGE 0.56), but integrating FDCs had substantial value. Because of the inaccurate representation of inputs, the baseline models might sometimes produce catastrophic results. However, model generalizability was further meaningfully improved by compiling an ensemble based on models with different input selections.

**Plain Language Summary**

There are large areas of land on all continents where we do not have streamflow measurements, but streamflow predictions are still needed under a changing climate. Machine learning models have recently demonstrated great success in modeling streamflow, but making predictions in ungauged regions (PUR) is still very risky, regardless of which type of model we use. Instead of requiring daily streamflow records used to train these models, there are other more available information sources such as the flow duration curve (FDC), which quantifies the long-term distribution of daily streamflow. We show that our deep learning model can assimilate information from the FDCs and present much more improved predictions for the PUR scenario. A neural network architecture is proposed for the integration such information. A novel combination of models with different input options was also found to improve the results. The results compare favorably with values reported in the literature. This is a significant step toward more reliable predictions in ungauged regions.

Main points:
1. An LSTM with an encoder can effectively assimilate flow duration curves (FDC) to improve streamflow predictions in ungauged regions.
2. The encoder unit extracted hydrologic features similar to baseflow index from FDC, and even sparsely-distributed FDC data can be helpful.
3. Input selection influences model robustness, while ensemble modeling with different inputs and FDC was effective in reducing risks.

---

[1] Corresponding author: cshen@engr.psu.edu

# 1. Introduction

Streamflow data is crucial for calibrating hydrologic models which quantify the water cycle (Wada et al., 2017) for various purposes ranging from climate modeling (Allen & Ingram, 2002) to climate change impact mitigation (Trabucco et al., 2008), from water sustainability studies to flood forecasting and humanitarian aid (Coughlan de Perez et al., 2016). Scanning over the Global Runoff Data Centre's worldwide map showing where streamflow data has been tracked (GRDC, 2020), one cannot help but notice vast swaths of lands with very few streamflow gauges, e.g., Asia, South America, Oceania, Central America, Africa, and even parts of the southwestern USA (Figure S1 in Supporting Information). In countries like China and Ethiopia, daily observations of streamflow are being recorded, but unfortunately not made openly accessible due to various reasons. In many cases, historical data (prior to the 1990s) are available, but it is difficult to obtain high-quality meteorological forcing data from the same periods. Many regions also typically lack data on physiographic attributes such as soil and aquifer properties. Such dearth of data causes huge gaps in our ability to calibrate and benchmark hydrologic models.

Prediction in ungauged basins (PUB) has been a longstanding topic in hydrology attracting prolonged interest (Hrachowitz et al., 2013), but this issue is still far from being solved. In a recent study, Beck et al. (2020) calibrated regionalized parameters (derived from different predictor maps) for 4,229 headwater catchments around the world and obtained a median Kling-Gupta efficiency (KGE) of 0.46 using randomized cross-validation, in comparison to 0.17 for the uncalibrated model. Although this is an improvement, we certainly desire better models. In comparison to randomized PUB, prediction in ungauged regions (PUR, no gauges or very sparse gauges in large regions. Here, *region* is intended to be a much larger spatial extent than *basins*) is even far riskier: with randomized PUB cross-validation, a certain geography and climate combination still has representation in the training (or calibration) dataset; with PUR, large swaths of lands with unique conditions could have no representation at all. PUR perhaps poses one of the most stringent tests on the generalizability of all kinds of hydrologic models.

While the gap in global daily streamflow data will continue, there are opportunities to obtain information about streamflow distribution, a relaxed form of information ("second-best") relevant to hydrologic processes. Hydrologic engineering handbooks have long published flow duration curves (FDC, a cumulative frequency curve that shows the percentiles of time for which specified discharge rates were equaled or exceeded during a given period). Regional FDCs have been provided for many data-scarce regions by computing FDCs from several basins in a region (Castellarin et al., 2004; Jian & Huan, 2009; M. Li et al., 2010; Masih et al., 2010; Mu et al., 2007; Nruthya & Srinivas, 2015; Smakhtin et al., 1997; Yu & Yang, 1996). Going forward, some satellite missions such as the imminent Surface Water and Ocean Topography (SWOT) mission have planned to provide monitoring of global streamflow for large rivers with a ~21 day revisit time (Biancamaria et al., 2016). After some years of sampling, we should be able to get better quantification of streamflow distributions.

Hydrologists have long employed FDCs to calibrate hydrologic models, but the main idea was to encourage the model to reproduce different aspects of the FDC such as distribution, volume, and slope, without forcing the model to match the daily streamflow exactly (Huan & Zhu, 2009; Shafii & Tolson, 2015; Westerberg et al., 2011). Characteristics extracted from the FDC were used as objective functions for calibration, which allowed the separation of runoff generation and routing processes. In this use case though, these models may not be optimal for flood forecasting in PUB or PUR, and need to be calibrated in individual basins, each with their respective FDCs.

Recently, the long short-term memory (LSTM) (Hochreiter & Schmidhuber, 1997) network, amongst other deep learning networks, has produced highly accurate streamflow predictions for both long-term forward runs and forecasting (Feng et al., 2020; Kratzert et al., 2018; W. Li et al., 2020; Shen, 2018; Xiang et al., 2020) where sufficient data are available. While time series DL models are rapidly developing momentum, to the best of our knowledge, no studies have examined the value of softer but more widely-available data such as FDCs to inform DL models.

While LSTM-based models have shown good performance, it would only be natural to think they may not be suitable for PUR due to their extensive data requirements. Kratzert et al. (2019) reported good PUB results using LSTM with extensive attributes, but testing was done only on neighboring random hold-out basins in the decently-gauged conterminous United States (CONUS) , with a very small holdout ratio for cross validation. Even so, some deterioration in LSTM performance was noted. DL models can be interpreted as Gaussian Processes, which make similar predictions for instances closer together in the input space (Fang et al., 2020; Gal & Ghahramani, 2016; Neal, 1996). Thus, in an extensively sampled geographic region, a DL model can make good predictions even if its internal representation for the inputs do not exactly capture fundamental physical relationships. However, if such representations are overfitted, the risk of giving catastrophic results for PUR increases dramatically. Gauch et al. (2020) demonstrated that in sparsely-distributed regions with shorter length of training data, the performance of an LSTM-based model could deteriorate rapidly. More attention needs to be paid towards mitigating such risks before predictions requiring spatial extrapolation can be considered trustworthy.

In this paper, we sought to test the performance and robustness of a LSTM-based streamflow model for large, contiguous regions with no gauges (PUR). We demonstrated, for the first time, the use of sparsely available FDCs to improve modeling, and the value of input-perturbed ensemble modeling. For hypothetical scenarios where FDCs were also only sparsely available, we tested if FDCs could be migrated from neighboring basins, as well as the impact of FDC availability at different densities. We also tested whether the input selection ensemble could reduce prediction risks, as the representations of inputs to LSTM are hypothesized to be a major source of this uncertainty.

# 2. Methods

## 2.1. Datasets

We ran our simulations on the Catchment Attributes and Meteorology for Large-sample Studies (CAMELS) dataset (Addor et al., 2017; Newman et al., 2014). CAMELS consists of basin-averaged hydrometeorological time series, catchment attributes, and streamflow observations for 671 reference catchments over the conterminous United States (CONUS). There are three different sources of meteorological forcing data in CAMELS; for this study, we used daily data from the North America Land Data Assimilation System (NLDAS). In addition to the 17 topographical, land cover, soil, and geological attributes used in our previous streamflow model (Feng et al., 2020), we also included 9 mean climate attributes (Table S1).

## 2.2. LSTM models with an encoder unit for FDC

We used a one-dimensional convolutional neural network (CNN), a deep learning unit referred to as our encoder unit, to extract static features from the input FDCs. The extracted features are concatenated with the original inputs of meteorological forcings and catchment attributes, and this combined information is then fed into the LSTM network to inform streamflow prediction. The combined CNN-LSTM framework (Figure S2) outputs the predicted discharge at each time step. For a more detailed discussion of the streamflow LSTM model, please refer to our previous study (Feng et al., 2020). Because the purpose of this network is to make predictions for ungauged regions, it does not use previous discharge observations as an input, unlike the setup in Feng et al. (2020). The hyperparameters for LSTM are the same as in our previous study. For the CNN encoder, we manually tuned the hyperparameters and chose optimal values (Table S2).

## 2.3. Experimental setup

We ran three types of experiments to measure model robustness given out-of-training prediction tasks. The temporal generalization test examined whether FDCs could provide new information for in-training basins (but for a different time period). The randomized PUB test examined whether the model could provide comparable results to previous benchmarks for cases of mild spatial extrapolation. With these two experiments providing context, the PUR tests (the core experiments of this study) were applied to show the value of FDCs and input ensemble modeling in data-scarce regions.

For temporal generalization and PUR tests, all models were trained using 10 years' worth of daily data from 10/01/1985 to 09/30/1995 and were tested from 10/01/1995 to 09/30/2005. To account for the stochasticity of DL models, we repeated six simulations with different random seeds for each experiment, and reported performance based on the mean of the predictions from these six members. For the randomized PUB test, we used the same setup as a previous benchmark study (Kratzert et al., 2019), which is further illustrated in section 2.3.1.

We trained models with different subsets of the static attributes as inputs. The full-attribute model setup (full-attr) used all available static attributes, while the no-attribute setup (no-attr) used no static basin attributes. Another setup used only 5 basic attributes (5-attr; catchment slope, area, forest fraction, soil porosity, and maximum soil water content). Averaging the discharge predictions from models with different input selections (3 input selections x 6 random seeds = 18 total simulations) returned an ensemble model we call the input selection ensemble. All the experiments were summarized in Table S3 in the supporting information and further illustrated below.

### 2.3.1. Temporal generalization and randomized PUB

We first trained LSTM-based streamflow models with and without FDCs on all the 671 CAMELS basins. The model without the FDC was our baseline, and other than the lack of lagged daily discharge observations, was the same as the one reported earlier (Feng et al., 2020). For each basin, the FDC was calculated as 100 discrete percentile values, based on the daily discharge observations available during the training period for that basin. For nearby PUB tests, we employed the same setup as Kratzert et al. (2019) to allow for direct comparison. As described earlier, their completely randomized PUB test did not hold out basins in a contiguous region, and is thus not an appropriate benchmark for the PUR experiment, but we nonetheless ran randomized holdout experiments with the same setup as theirs to make sure our LSTM network had similar performance. For this test, we used k-fold (k=12) cross validation, in which the 531 CAMELS basin subset (which excluded basins with uncertain boundaries or very large areas) were randomly divided into 12 groups with approximately equal overall sizes. The experiments were executed 12 times, with each group in turn serving as the test set while the rest of the basins comprised the training set, in order to cover all 531 basins. Furthermore, we examined the effects of using FDCs for the test basins or migrating the nearest FDC information gauged basins to them.

### 2.3.2. Prediction in ungauged regions (PUR)

To test the performance of our models in scenarios of PUR, we divided the whole CONUS into seven contiguous regions based on the spatial distribution of the CAMELS basins and the first level classification of Hydrologic Unit Maps (HUC2), as shown in Figure S3 and Table S4 in the supporting information. This rough division into contiguous regions, as opposed to division by physiographic provinces or biomes, is in agreement with the purpose of this test; as in real-world applications, data availability often does not agree with physiographic divisions. Since the numbers of CAMELS basins within each HUC2 region were quite different, we combined neighboring HUC regions into larger regions, and tried to keep the number of basins in each region similar. The resulting 7 large regions have an average area of 12.6x$10^5$ km$^2$, and we refer to each of these as "PUR regions". We note the gages are quite sparse, ranging from 4.4 to 21 gauges per $10^5$ km$^2$. Similar to the random hold-out experiments previously described, streamflow models were trained on the basins in six of the seven regions, and model performance tested in the remaining held-out region.

We set up different PUR test scenarios with the available FDC information. In one test scenario, FDCs were considered to be available for all the basins in the target ungauged region. However, we also tested scenarios where only a fraction (1/3 or 1/10) of the basins in a target region had FDCs available. For a basin lacking its own FDC, the input to the CNN was the FDC from the nearest basin with this information available. These tests were to examine the potential to spatially extend sparse FDC information in the region. It also mimics the situation where regional FDCs were constructed from multiple gauges in a region. If only ⅓ of the FDCs were considered to be available, the density of FDCs ranged from 0.4 to 2.1 gauges per $10^5$ km$^2$, which is a low information density (Table S4).

# 3. Results and Discussions

## 3.1. Temporal generalization and randomized PUB tests

In the densely sampled temporal generalization experiment, the CNN encoder was able to extract features from the FDC to partially compensate for the missing information when static input attributes were not provided, but it could not elevate the model's performance if attributes were provided (Figure 1). Static attributes were shown to be crucial inputs to the streamflow model trained on all CAMELS basins, as the median NSE value deteriorated from 0.74 to 0.58 when attributes were withheld. The model integrating FDCs significantly recovered the performance lost by withholding attributes, resulting in an NSE value of 0.66.

In contrast, the inclusion of FDC information alongside the attributes were essentially equivalent to the scenario with attributes but without FDCs. This observation suggested that, given this data density and the experimental setup, the information contained in FDCs was redundant to and encompassed by the information contained in the input attributes and the observed discharge (provided as the target variable) during training.

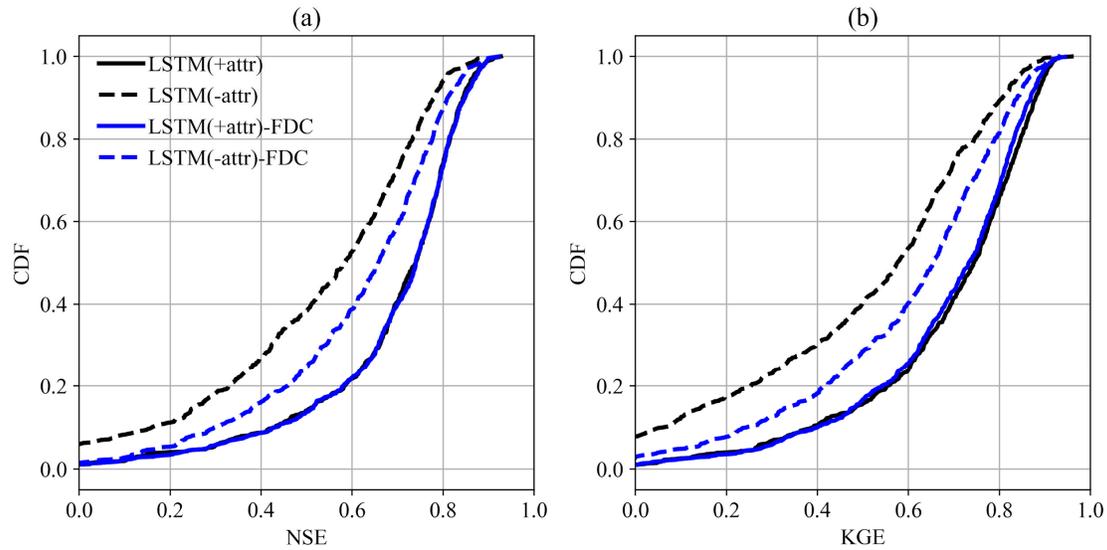

Figure 1. Temporal generalization test: the performance of models with or without the CNN FDC encoder or input basin attributes when trained on all CAMELS basins. "+attr" or "-attr" in the legend indicates whether basin attributes were or were not included as inputs, respectively.

For the randomized PUB test, we acquired NSE median values that were essentially equivalent to those in the previous benchmark study (Kratzert et al., 2019), showing the validity of our model setup (Text S1, Figure S4 in the supporting information). Since our model performed very closely as theirs, the decline in the PUR results in the next section can be attributed to the inherent risk of PUR. Similar to the temporal generalization test, integrating FDCs from nearby basins significantly improved predictions for the model without attributes but only marginally improved the model with full input attributes. Please see the supporting information for more detailed discussions on the randomized PUB test.

## 3.2. Prediction in ungauged regions (PUR)

For our core PUR tests over the CONUS, the input-selection ensemble models with no FDCs, all FDCs, 1/3 FDCs, and 1/10 FDCs obtained median NSE values of 0.568, 0.614, 0.605, and 0.597, respectively (Figure 2h). The corresponding KGE values were 0.556, 0.619, 0.601, and 0.585, respectively (Figure S5 in the Supporting Information). All of these values, even for models without using FDCs, appear favorable in the light of the best global PUB literature values, which had a median KGE of ~0.46 (Beck et al., 2020). Both Kratzert et al. (2019) and Beck et al. (2020) are PUB rather than PUR, but we referenced Beck et al., (2020)'s value here because, although it is also completely randomized holdout, they had a larger holdout ratio in a more sparse dataset, and is a relatively more appropriate reference point for the our PUR tests. Their values are not directly comparable due to the different datasets and holdout strategies employed, but nonetheless highlight our success, especially considering that (i) PUR should be more challenging than the randomized PUB reported in Beck et al. (2020) and we expect the model results in the literature to further deteriorate if PUR was tested; (ii) the forcing data used in the global PUB studies were of high quality and were not the reason of lowered performance based on their local calibration results (Beck et al., 2020); and (iii) values for our randomized PUB tests were much higher (Figure S4). By adding FDCs and input-perturbed ensemble modeling, we obtained the strongest PUR models reported to date.

Despite relatively decent performance in all test regions and for all input selections (Figure 2), we noticed significant performance deterioration from the reference model trained on all CAMELS basins (the first red box) to the baseline PUR models (the second "No FDC" group). For the all-attribute model, the median NSE decreased in the range of 0.10 to 0.60, with an average of 0.22, which highlights the risks of PUR.

Among these PUR tests, Region 6 (Southwestern US) stands out: whereas the in-training reference (the first red box in Figure 2f) had a median NSE of 0.74, the highest value achieved by the four PUR models with the all-attribute input was 0.17. This huge gap not only suggests that Region 6 has different hydrologic responses in runoff generation than the other CONUS regions, but also implies the PUR models trained on other regions (especially with all input attributes) did not capture the required fundamental hydrologic principles. If a relationship were universal, it should be universally applicable and should not see such declines. Region 6, except near the coast, is characterized by an arid environment dominated by groundwater flow. Presumably, geology has a strong impact in this region, but was not learned correctly in the model trained on other regions. The catastrophic results for the baseline (no FDC) PUR models were not limited to Region 6 or the full-attribute model, however. The 5-attribute baseline PUR model gave a large spread in NSE and the median dropped below 0.3 in Region 7 (northwestern US) (Figure 2g), which is characterized by the presence of the Rocky Mountains, and large precipitation gradients. In most other regions, the no-attribute model had relatively poor performance.

These observations confirmed that PUR models indeed incur substantial risks and should not be applied blindly. One might expect that models with fewer inputs could have a higher chance of learning the true underlying dynamics related to these input variables, and thus less risk of overfitting to noise in the inputs. Performance results in Region 6 seemed to support this theory, as the no-attribute and 5-attribute models performed better than the full-attribute one, but Region 7 results countered this trend. It is possible that some of the five attributes were still overfitted. Thus it is challenging to anticipate when the model would deteriorate badly.

The input selection ensemble greatly enhanced the results in 5 of the 7 regions compared to the best member models, raising NSE values by an average of 0.032 from the baseline (No FDC) full-attribute model. When plotted, the lower whiskers of the ensemble tended to be higher than the individual models that comprise the ensemble, suggesting the ensemble model is more robust. More importantly, the input-selection ensemble encountered no aberrantly bad results as the member models sometimes did.

Among the ensemble members, we noticed that no single input options always gave the best model results, and there were no fixed performance orders among the members. While results do follow the order $NSE_{no\text{-}attr} < NSE_{5\text{-}attr}$ or $NSE_{full\text{-}attr} < NSE_{ens}$ (the order between $NSE_{full\text{-}attr}$ and $NSE_{5\text{-}attr}$ is fluctuating) for 5 of the regions, Regions 6 and 7 were exceptions. This is a clear sign that the full-attribute model incorrectly modeled the effects of the input attributes, which then became a major source of extrapolation error, despite this model scoring great metrics in the randomized neighbor test (Figure S4). Models with different input selections forced different representations, and thus their ensemble mean cancelled out some of the errors and effectively reduced true risks.

Assimilating FDCs elevated model performances in 6 of the 7 regions but had little impact in Region 4. Of the 6 regions where FDCs were beneficial, the information from FDCs improved almost all the input selections and the ensemble. It improved the median $NSE_{full\text{-}attr}$, $NSE_{no\text{-}attr}$, $NSE_{5\text{-}attr}$, and $NSE_{ens}$ by an average of 0.033, 0.079, 0.083, and 0.051, respectively. It seems the impacts of FDCs were more diluted for the full-attribute model because this model was trained to place more importance on the attributes rather than the FDCs.

The value of FDCs degraded gradually as the availability of FDCs decreased from the full target region to sparse (1/3 basins considered FDC available) and very low (1/10 basins considered FDC available) densities. However, for the ensemble model, the impacts of the FDCs were obvious even at very low availability density (1/10 FDCs) for 4 of the 7 regions (Regions 1, 2, 6, and 7). This result seemed to suggest the features extracted by the FDC had long-ranged spatial autocorrelation.

So, what features did the CNN unit extract from the FDCs? Since the network conditioned the CNN unit to optimize streamflow prediction, the extract features could also reveal what information was important for that purpose. Figure 3 plots the spatial maps of three CNN features from the no-attribute CAMELS model, along with maps of several attributes. There are distinct clustering of high- and low-value regions in feature A (Figure 3a). It seems roughly inline with the difficult-to-model basins with low NSE values (Figure 3i). For feature B (Figure 3b), the low values concentrate on the mountainous and snow-dominant regions (see the maps of slope and fraction of snow in Figure 3e-f), such as the Appalachian range, New England, the Great Lakes, the Rocky Mountains, and northwest coast ranges. Most of these regions have the streamflow characteristic of a high autocorrelation function (Figure 3h). For the semi-arid basins on the great plains in the central CONUS, feature B tended to have higher values. Without the instruction to do so, one feature well mimicked the pattern of baseflow index, as shown by comparing feature C (Figure 3c) with the baseflow index map (Figure 3g). This suggests the CNN encoder can learn to calculate the low-flow portion of the total flows from the FDC information. It also suggests that this baseflow ratio is relevant to the correct estimation of the hydrographs. Our previous study identified LSTM prediction of baseflow in the western US as a challenge, because of the limitations of the inputs related to geology (Feng et al., 2020). Furthermore, since baseflow is closely related to groundwater processes, we expect that this encoder may be capable of inferring some geological characteristics from the FDC information as well.

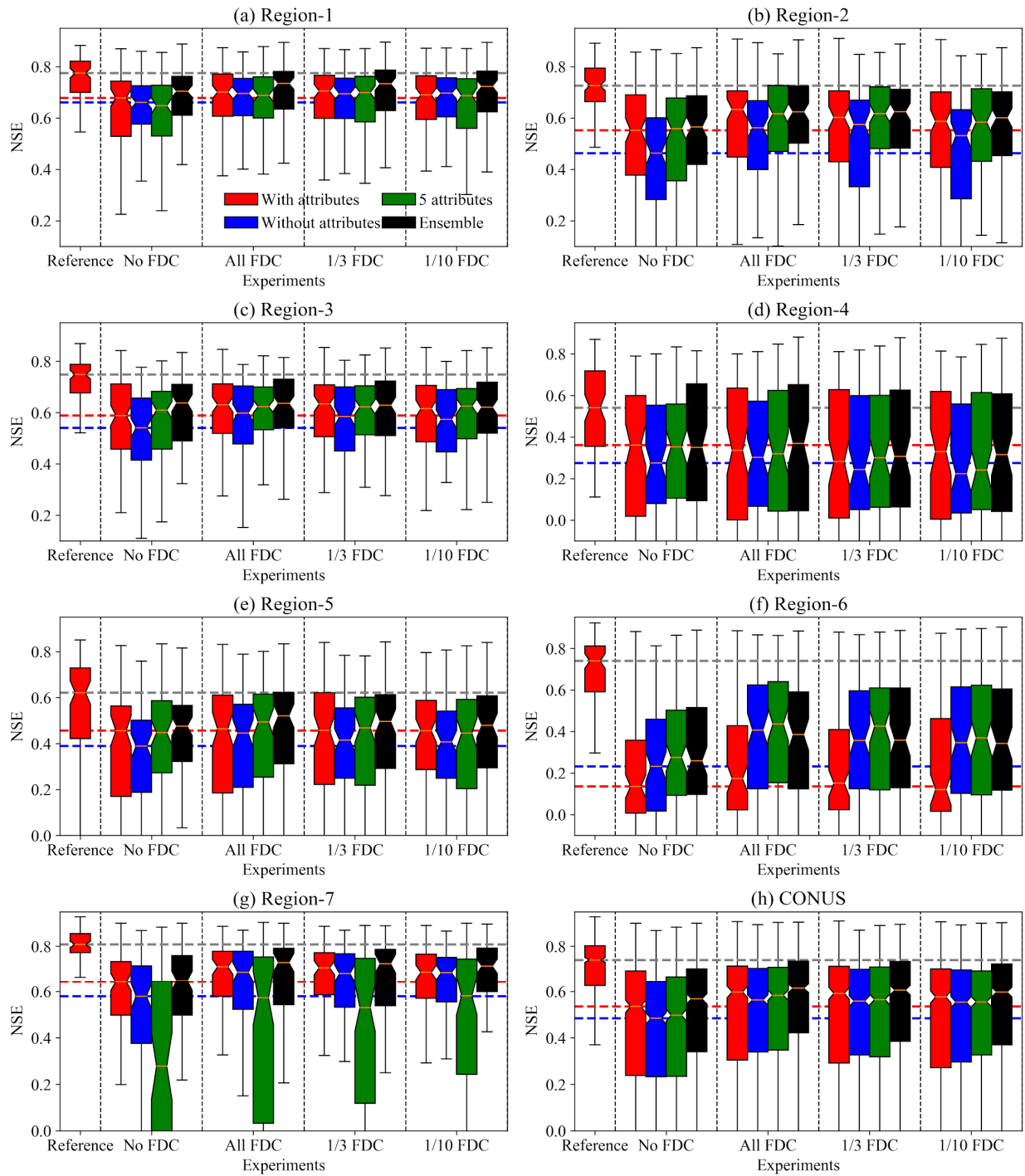

Figure 2. The NSE of PUR tests for different regions (a-g) as the holdout sets and (h) whole CONUS from regional cross-validation. For the first reference solution, the model was trained with the whole CAMELS dataset, i.e., no basins were held out. ⅓ and 1/10 FDC mean only ⅓ and 1/10 of the test basins had FDCs and they were migrated to the nearest basins without FDCs.

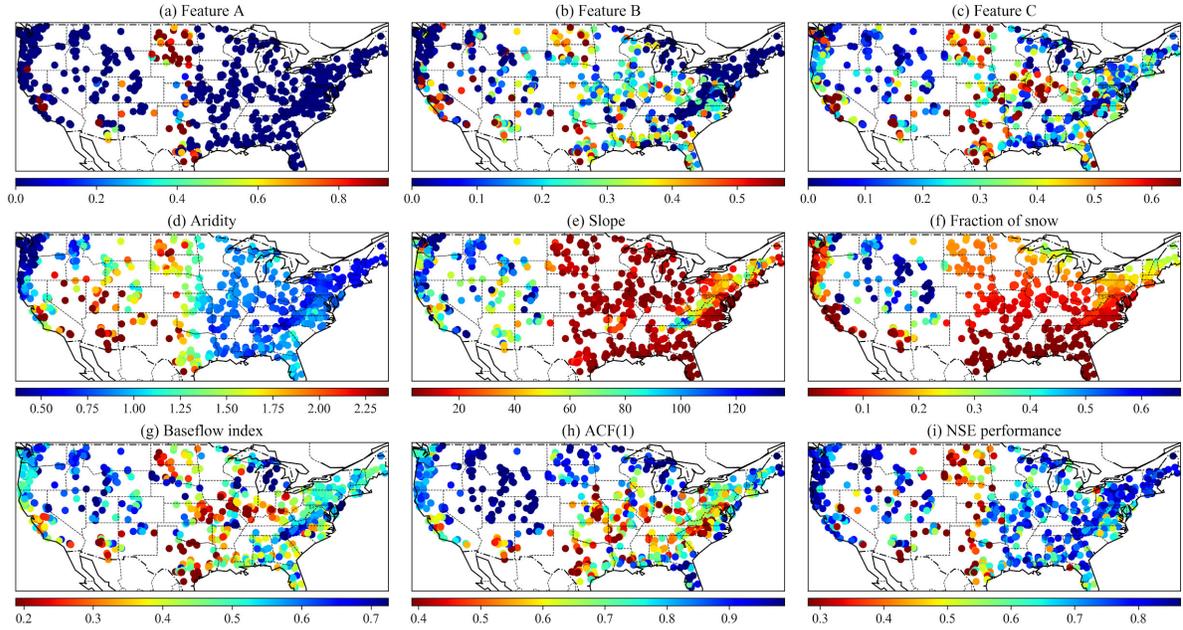

*Figure 3. Maps of the features extracted by the CNN encoder from the no-attribute model and the basin characteristics. (a)-(c) Three selected features; (d) Aridity: ratio between annual potential evapotranspiration and precipitation; (e) Slope: basin mean slope; (f) Fraction of snow: the fraction of precipitation falling as snow; (g) Baseflow index; (h) ACF(1): 1-day-lag autocorrelation function of streamflow; (i) Full-attribute model performance evaluated by NSE value.*

## 3.4. Further discussion

How do we interpret why the input selection ensemble worked? The CNN encoder created some intermediate embedded features (x'). We may regard that x' in fact contains two parts of information: catchment attributes (a') such as slope which overlap with the static physiographic input attributes (A), as well as flow-related attributes (q') such as ranges of flow which do not directly overlap with A:

$[a', q'] = x' = CNN(FDC)$

Then, they are used in combination with A and other inputs to predict discharge:

$Q_{mod}^{1:t} = LSTM(a', q', A, F^{1:t})$

where $Q_{mod}$ represents the modeled discharge, t is the length of the training time series, and F represents the forcing variables.

In our interpretation of the results, the model put different weights on [a', q', A] as modulating factors of the dynamical model component. When A was extensive, as in the full-attribute model, the strongest weights were placed on A. When no A was provided (no-attribute model), the model was forced to entirely rely on [a', q'] to model the spatial heterogeneity in characteristics. When A was limited, as in the 5-attribute model, the weights were split more evenly between A, a', and q'. Whenever A was involved in a significant way, there was a risk of overfitting because there may not have been sufficient combinations of A in the training data to allow the model to fully resolve the different impacts of these variables. While FDCs only contain limited information (for example, they do not contain information related to timing or matched forcings and responses), q' has a more uniform relationship everywhere. This means that the no-attribute, with-FDC model was more robust, but not strong enough to make up for the missing information contained in A. In extensively sampled regions, because daily streamflow appears in the target data, the model can be trained to infer information similar to q' so q' do not provide noticeable benefits. However, when ported in a new region in PUR, the information in q' serves as a stabilizing factor to make the models more robust.

The results suggest, in an extrapolation setting, it is not always better for data-driven models to have more input attributes for instances that may be very different from the training set. Under the premise of limited data, increasing input attributes increases the variance (in the context of the bias-variance tradeoff) of the model and the risk of overfitting. We should be cautious when trying to interpret model representations or embeddings, especially in a small-data setting. Using an input selection ensemble can reduce the true risk of the model given the available training data, but it still cannot inform us of the unobservable processes in the ungauged basins. Hence, the PUR model performance gives us a lower-bound estimate of the commonality in the hydrologic processes between the training and the PUR test regions. Also, as noted earlier, the model performance seems related to features extracted by the FDC (Figure 3i vs. 3b), which suggests we can use the network to model uncertainty terms, in agreement with our earlier study (Fang et al., 2020).

# 4. Conclusion

Prediction in ungauged regions presents perhaps one of the toughest challenges to hydrologic modeling and flood forecasting. Our results suggest that even without FDC information, LSTM models can produce competitive predictions for PUR. Given FDC information and an input selection ensemble, LSTM models can reach a median PUR NSE of 0.61 (median KGE = 0.62) over the CONUS, which is markedly higher than the best results reported in the literature (~0.45 median KGE). FDC information was thus valuable, and could be utilized through an encoder. More than half of the basins in the conterminous United States now have what can be considered to be a functional model. In the future, we may also be able to produce uncertainty estimates that advise us of risks with PUR.

The use of an input selection ensemble was highly useful because LSTM models have varied internal representations for the inputs, and ensemble averaging reduces the risk of overfitting. The CNN structure was effective at extracting features from FDCs to further improve model performance. Some of the extracted features resembled baseflow and other features related to factors which were not adequately described by the physiographic inputs. In the future where satellite missions can provide a better description of streamflow distribution, along with other hydrologic variables like soil moisture (Fang et al., 2017; Fang & Shen, 2020), we will likely be able to make even better predictions for the dreaded PUR problem.

# 5. Acknowledgments

DF and CS were supported by US National Science Foundation Award EAR#1832294. Data for CAMELS can be downloaded at https://ral.ucar.edu/solutions/products/camels.